\documentstyle[aclap1]{article}

\input psfig

\newcommand{\drt}[1]{\langle#1\rangle}

\newcommand{\ignore}[1]{}

\newcommand{\phantomhack}{\phantom{haunt haunt haunt haunt haunt haunt haunt!}}
\newcommand{\mylefteqn}[1]{\lefteqn{#1} & & \phantomhack}

\title{\vspace{-0.5in}Approximating Context-Free Grammars \\ with a Finite-State Calculus}
\author{Edmund GRIMLEY EVANS\\
Computer Laboratory\\
University of Cambridge\\
Cambridge, CB2 3QG, GB\\
{\tt Edmund.Grimley-Evans@cl.cam.ac.uk}}

\begin{document}
\bibliographystyle{fullname}
\maketitle
\vspace{-0.5in}
\begin{abstract}
Although adequate models of human language for syntactic analysis and
semantic interpretation are of at least context-free complexity, for
applications such as speech processing in which speed is important
finite-state models are often preferred. These requirements may be
reconciled by using the more complex grammar to automatically derive a
finite-state approximation which can then be used as a filter to guide
speech recognition or to reject many hypotheses at an early stage of
processing. A method is presented here for calculating such
finite-state approximations from context-free grammars. It is
essentially different from the algorithm introduced by Pereira and
Wright \shortcite{Pereira91,Pereira96}, is faster in some cases, and has the
advantage of being open-ended and adaptable.
\end{abstract}

\section{Finite-state approximations}

Adequate models of human language for syntactic analysis and semantic
interpretation are typically of context-free complexity or
beyond. Indeed, Prolog-style definite clause grammars (DCGs) and
formalisms such as PATR with feature-structures and unification have
the power of Turing machines to recognise arbitrary recursively
enumerable sets. Since recognition and analysis using such models may
be computationally expensive, for applications such as speech
processing in which speed is important finite-state models are often
preferred.

When natural language processing and speech recognition are integrated
into a single system one may have the situation of a finite-state
language model being used to guide speech recognition while a
unification-based formalism is used for subsequent processing of the
same sentences. Rather than write these two grammars separately, which
is likely to lead to problems in maintaining consistency, it would be
preferable to derive the finite-state grammar automatically from the
(unification-based) analysis grammar.

The finite-state grammar derived in this way can not in general
recognise the same language as the more powerful grammar used for
analysis, but, since it is being used as a front-end or filter, one
would like it not to reject any string that is accepted by the analysis
grammar, so we are primarily interested in `sound approximations' or
`approximations from above'.

Attention is restricted here to approximations of context-free grammars
because context-free languages are the smallest class of formal
language that can realistically be applied to the analysis of natural
language. Techniques such as restriction \cite{ShieberRes} can be used to construct
context-free approximations of many unification-based formalisms, so
techniques for constructing finite-state approximations of context-free
grammars can then be applied to these formalisms too.

\section{Finite-state calculus}

A `finite-state calculus' or `finite automata toolkit' is a set of
programs for manipulating finite-state automata and the regular
languages and transducers that they describe. Standard
operations include intersection, union, difference,
determinisation and minimisation. Recently a number of automata
toolkits have been made publicly available, such as FIRE Lite
\cite{Watson96}, Grail \cite{Grail96}, and FSA Utilities
\cite{vanNoord96}.

Finite-state calculus has been successfully applied both to morphology
\cite{Kaplan:94,Kempe96} and to syntax
(constraint grammar, finite-state syntax).

The work described here used a finite-state calculus implemented by the
author in SICStus Prolog. The use of Prolog rather than C or C++ causes
large overheads in the memory and time required. However, careful
account has been taken of the way Prolog operates, its indexing in
particular, in order to ensure that the asymptotic complexity is
as good as that of the best published algorithms, with the result that
for large problems the Prolog implementation outperforms some of the
publicly available implementations in C++. Some versions of the calculus
allow transitions to be labelled with arbitrary Prolog terms, including
variables, a feature that proved to be very convenient for prototyping
although it does not essentially alter the power of the machinery. (It
is assumed that the string being tested consists of ground terms so no
unification is performed, just matching.)

\section{An approximation algorithm}\label{algsect}

There are two main ideas behind this algorithm. The first is to
describe the finite-state approximation using formulae with regular
languages and finite-state operations and to evaluate the formulae
directly using the finite-state calculus. The second is to use, in
intermediate stages of the calculation, additional, auxiliary symbols
which do not appear in the final result. A similar approach has been used for
compiling a two-level formalism for morphology \cite{Grimley96}.

In this case the auxiliary symbols are dotted rules from the given
context-free grammar.
A dotted rule is a grammar rule with a dot inserted somewhere
on the right-hand side, e.g.
\begin{verse}
S $\rightarrow$ $\cdot$ NP VP\\
S $\rightarrow$ NP $\cdot$ VP\\
S $\rightarrow$ NP VP $\cdot$
\end{verse}

However, since these dotted rules are to be used as terminal symbols of
a regular language, it is convenient to use a more compact notation:
they can be replaced by a triple made out of the nonterminal symbol on
the left-hand side, an integer to determine one of the productions
for that nonterminal, and an integer to denote the
position of the dot on the right-hand side by counting the
number of symbols to the left of the dot. So, if `S $\rightarrow$ NP
VP' is the fourth production for S, the dotted rules given
above may be denoted by $\langle S,4,0\rangle$, $\langle S,4,1\rangle$
and $\langle S,4,2\rangle$, respectively.

It will turn out to be convenient to use a slightly more complicated
notation: when the dot is located after the last symbol on the
right-hand side we use $z$ as the third element of the triple instead
of the corresponding integer, so the last triple is $\langle
S,4,z\rangle$ instead of $\langle S,4,2\rangle$. (Note that $z$ is an
additional symbol, not a variable.) Moreover, for epsilon-rules, where
there are no symbols on the right-hand side, we treat the $\epsilon$ as
it were a real symbol and consider there to be two corresponding dotted
rules, e.g.\ $\langle MOD,1,0\rangle$ and $\langle MOD,1,z\rangle$
corresponding to `MOD $\rightarrow$ $\cdot$ $\epsilon$' and `MOD
$\rightarrow$ $\epsilon$ $\cdot$' for the rule `MOD $\rightarrow$
$\epsilon$'.

Using these dotted rules as auxiliary symbols we can work with regular
languages over the alphabet
\begin{eqnarray*}
\mylefteqn{\Sigma = T \cup \{\,\langle X,m,n\rangle
\mid X \in V \wedge m=1,\ldots,m_X \wedge} \\ & &
n=0,\ldots,\max\{n_{X,m}-1,0\},z\,\}
\end{eqnarray*}
where $T$ is the set of terminal symbols, $V$ is the set of
nonterminals, $m_X$ is the number of productions for nonterminal $X$,
and $n_{X,m}$ is the number of symbols on the right-hand side of the
$m$th production for $X$.

It will be convenient to use the symbol $*$ as a `wildcard', so
$\langle s,*,0\rangle$ means $\{\,\langle X,m,n\rangle \in \Sigma \mid
X=s, n=0\,\}$ and $\langle *,*,z\rangle$ means $\{\,\langle
X,m,n\rangle \in \Sigma \mid n=z\,\}$. (This last example explains why
we use $z$ rather than $n_{X,m}$; it would otherwise not be possible to
use the `wildcard' notation to denote concisely the set $\{\,\langle
X,m,n\rangle \mid n=n_{X,m}\,\}$.)

We can now attempt to derive an expression for the set of strings over
$\Sigma$ that represent a valid parse tree for the given grammar: the
tree is traversed in a top-down left-to-right fashion and the daughters
of a node X expanded with the $m$th production for X are separated by
the symbols $\drt{X,m,*}$. (Equivalently, one can imagine the auxiliary
symbols inserted in the appropriate places in the right-hand side of
each production so that the grammar is then unambiguous.) Consider, for
example, the following grammar:
\begin{verse}
S $\rightarrow$ a S b\\
S $\rightarrow$ $\epsilon$
\end{verse}
Then the following is one of the strings over $\Sigma$ that we would like to
accept, corresponding to the string $aabb$ accepted by the grammar:
\begin{eqnarray*}
\mylefteqn{\langle s,1,0\rangle a \langle s,1,1\rangle
\langle s,1,0\rangle a \langle s,1,1\rangle
\langle s,2,0\rangle \langle s,2,z\rangle} \\ & &
\langle s,1,2\rangle b \langle s,1,z\rangle
\langle s,1,2\rangle b \langle s,1,z\rangle
\end{eqnarray*}

Our first approximation to the set of acceptable strings is
$\drt{S,*,0} \Sigma^* \drt{S,*,z}$, i.e.\ strings that start with
beginning to parse an S and end with having parsed an S. From this
initial approximation we subtract (that is, we intersect with the
complement of) a series of expressions representing restrictions on the
set of acceptable strings:\footnote {In these expressions over regular
languages set union and set difference are denoted by $+$ and $-$,
respectively, while juxtaposition denotes concatenation and the bar
denotes complementation ($\overline{x}\equiv\Sigma^*-x$).}


\begin{equation}\label{e1}
\overline{(\Sigma^*(\drt{*,*,*}-\drt{*,*,z}))+\epsilon}\drt{*,*,0}\Sigma^*
\end{equation}

Formula \ref{e1} expresses the restriction that a dotted rule of the
form $\drt{*,*,0}$, which represents starting to parse the right-hand
side of a rule, may be preceded only by nothing (the start of the
string) or by a dotted rule that is not of the form $\drt{*,*,z}$
(which would represent the end of parsing the right-hand side of a
rule).

\begin{equation}\label{e2}
\Sigma^*\drt{*,*,z}\overline{\epsilon+(\drt{*,*,*}-\drt{*,*,0})\Sigma^*}
\end{equation}

Formula \ref{e2} similarly expresses the restriction that a dotted rule
of the form $\drt{*,*,z}$ may be followed only by nothing
or by a dotted rule that is not of the form $\drt{*,*,0}$.

For each non-epsilon-rule with 
dotted rules $\drt{X,m,n}$,
$n=0,\ldots,n_{X,m}-1,z$, for each $n=0,\ldots,n_{X,m}-1$:
\begin{equation}\label{e3}
   \Sigma^*\drt{X,m,n}\overline{{\rm next}(X,m,n+1)\Sigma^*}
\end{equation}
where
\begin{eqnarray*}
\lefteqn{{\rm next}(X,m,n)=} \\
 & a\drt{X,m,n} & ({\rm rhs}(X,m,n)=a,\; a\in T,\; n<n_{X,m}) \\
 & a\drt{X,m,z} & ({\rm rhs}(X,m,n)=a,\; a\in T,\; n=n_{X,m}) \\
 & \drt{A,*,0}  & ({\rm rhs}(X,m,n)=A,\; A\in V)
\end{eqnarray*}
where ${\rm rhs}(X,m,n)$ is the $n$th symbol on the right-hand side of
the $m$th production for $X$.

Formula \ref{e3} states that the dotted rule $\drt{X,m,n}$ must be followed by $a\drt{X,m,n+1}$ (or
$a\drt{X,m,z}$ when $n+1=n_{X,m}$) when the next item to be parsed is the
terminal $a$, or by $\drt{A,*,0}$ (starting to parse an $A$) when the
next item is the nonterminal $A$.

For each non-epsilon-rule with 
dotted rules $\drt{X,m,n}$,
$n=0,\ldots,n_{X,m}-1,z$, for each $n=1,\ldots,n_{X,m}-1,z$:
\begin{equation}\label{e4}
   \overline{\Sigma^*{\rm prev}(X,m,n)}\drt{X,m,n}\Sigma^*
\end{equation}
where
\begin{tabbing}
${\rm prev}(X,m,n)=$ \\
  $\drt{X,m,n_{X,m}-1}a$ \=\kill
  $\drt{X,m,n-1}a$       \> $({\rm rhs}(X,m,n)=a,\; a\in T,\; n\not=z)$ \\
  $\drt{X,m,n_{X,m}-1}a$ \> $({\rm rhs}(X,m,n)=a,\; a\in T,\; n=z)$ \\
  $\drt{A,*,z}$          \> $({\rm rhs}(X,m,n)=A,\; A\in V)$
\end{tabbing}

Formula \ref{e4} similarly states that the dotted rule $\drt{X,m,n}$ must be preceded by $\drt{X,m,n-1}a$ (or
$\drt{X,m,n_{X,m}-1}$ when $n=z$) when the previous item was the
terminal $a$, or by $\drt{A,*,z}$ when the
previous item was the nonterminal $A$.

For each epsilon-rule corresponding to dotted rules $\drt{X,m,0}$ and
$\drt{X,m,z}$:
\begin{equation}\label{e5}
   \Sigma^*\drt{X,m,0}\overline{\drt{X,m,z}\Sigma^*}
\mbox{, and}
\end{equation}
\begin{equation}\label{e6}
   \overline{\Sigma^*\drt{X,m,0}}\drt{X,m,z}\Sigma^*
\end{equation}

Formulae \ref{e5} and \ref{e6} state that the dotted rule $\drt{X,m,0}$
must be followed by $\drt{X,m,z}$, and $\drt{X,m,z}$ must be preceded
by $\drt{X,m,0}$.

For each non-epsilon rule with dotted rules $\drt{X,m,n}$,
$n=0,\ldots,n_{X,m}-1,z$, for each $n=0,\ldots,n_{X,m}-1$:
\begin{equation}\label{e7}
   \Sigma^*\!\drt{X,m,n}
   \overline{(\Sigma\!-\!\drt{X,m,*})^*(\drt{X,m,0}\!+\!\drt{X,m,n'})\Sigma^*}
\end{equation}
and
\begin{equation}\label{e8}
   \overline{\Sigma^*\!(\drt{X,m,z}\!+\!\drt{X,m,n})(\Sigma\!-\!\drt{X,m,*})^*}
   \drt{X,m,n'}\Sigma^*
\end{equation}
where
\[ n'=\cases{
   n+1,& if $n<n_{X,m}-1$;\cr
   z,& if $n=n_{X,m}-1$.\cr} \]

Formula \ref{e7} states that the next instance of $\drt{X,m,*}$ that
follows $\drt{X,m,n}$ must be either $\drt{X,m,0}$ (a recursive
application of the same rule) or $\drt{X,m,n'}$ (the next stage in
parsing the same rule), and there must be such an instance. Formula
\ref{e8} states similarly that the closest instance of $\drt{X,m,*}$
that precedes $\drt{X,m,n'}$ must be either $\drt{X,m,z}$ (a recursive
application of the same rule) or $\drt{X,m,n}$ (the previous stage in
parsing the same rule), and there must be such an instance.


When each of these sets has been subtracted from the initial
approximation we can remove the auxiliary symbols (by applying the
regular operator that replaces them with $\epsilon$) to give the final
finite-state approximation to the context-free grammar.

\section{A small example}

It may be admitted that the notation used for the dotted rules was
partly motivated by the possibility of immediately testing the
algorithm using the finite-state calculus in Prolog: the regular
expressions listed above can be evaluated directly using the `wildcard'
capabilities of the finite-state calculus.

Figure~\ref{codefig} shows the sequence of calculations that corresponds to
applying the algorithm to the following grammar:
\begin{verse}
S $\rightarrow$ a S b \\
S $\rightarrow$ $\epsilon$
\end{verse}
With the following notational explanations it should be possible to
understand the code and compare it with the description of the
algorithm.
\begin{itemize}

\item
The procedure \verb|r(RE,X)| evaluates the regular expression \verb|RE|
and puts the resulting (minimised) automaton into a register with the
name \verb|X|.

\item
\verb|list_fsa(X)| prints out the transition table for the
automaton in register \verb|X|.

\item
Terminal symbols may be any Prolog terms, so the terminal alphabet is
implicit. Here atoms are used for the terminal symbols of the grammar
(\verb|a| and \verb|b|) and terms of the form \verb|_/_/_| are used for
the triples representing dotted rules. The terms need not be ground, so
the Prolog variable symbol \verb|_| is used instead of the `wildcard'
symbol $*$ in the description of the algorithm.

\item
In a regular expression:
\begin{itemize}

\item \verb|#X| refers to the contents of register \verb|X|;

\item \verb|$| represents $\Sigma$, any single terminal symbol;

\item  \verb|s| represents a string of terminals with length equal to the
number of arguments; so \verb|s| with no arguments represents the empty string
$\epsilon$, \verb|s(a)| represents the single terminal $a$, and
\verb|s(s/_/0)| represents the dotted rules $\drt{s,*,0}$;

\item Kleene star is \verb|*| (redefined as a postfix operator), and
concatenation and union are \verb|^| and \verb|+|, respectively;

\item other operators provided include \verb|&| (intersection) and
\verb|-| (difference); there is no operator for complementation;
instead subtraction from $\Sigma^*$ may be used, e.g.\ \verb|($ *)-(#l)|
instead of $\overline{L}$;

\item \verb|rem(RE,L)| denotes the result of removing from the language
\verb|RE| all terminals that match one of the expressions in the list
\verb|L|.

\end{itemize}
\end{itemize}

\ignore{
i(1). t(1,det,2). t(1,prep,1). t(1,v,3).
t(2,adjp,2). t(2,n,1).
t(3,det,4). t(3,prep,3). t(3,v,5).
t(4,adjp,4). t(4,n,6).
t(5,conj,7). t(5,det,8). t(5,prep,5). t(5,v,5).
t(6,det,9). t(6,prep,3). t(6,v,5).
t(7,conj,10). t(7,det,11). t(7,prep,10). t(7,v,5).
t(8,adjp,8). t(8,n,12).
t(9,adjp,9). t(9,n,13).
t(10,conj,10). t(10,det,11). t(10,prep,10). t(10,v,14).
t(11,adjp,11). t(11,n,10).
t(12,conj,7). t(12,det,11). t(12,prep,5). t(12,v,5).
t(13,det,9). t(13,prep,13). t(13,v,14).
t(14,conj,10). t(14,det,15). t(14,prep,14). t(14,v,5).
t(15,adjp,15). t(15,n,16).
t(16,conj,10). t(16,det,11). t(16,prep,14). t(16,v,5).
f(5).
f(12).
}

The context-free language recognised by the original context-free
grammar is $\{\,a^nb^n \mid n \ge 0\,\}$. The result of applying the
approximation algorithm is a 3-state automaton recognising the language
$\epsilon+a^{+}b^{+}$.

\section{Computational complexity}
\label{sectcompcomp}

Applying the restrictions expressed by formulae \ref{e1}--\ref{e6}
gives an automaton whose size is at most a small constant
multiple of the size of the input grammar. This is because these
restrictions apply locally: the state that the automaton is in after
reading a dotted rule is a function of that dotted rule.

When restrictions \ref{e7}--\ref{e8} are applied the final automaton
may have size exponential in the size of the input grammar. For
example, exponential behaviour is exhibited by the following class of
grammars:
\begin{verse}
S $\rightarrow$ a$_1$ S a$_1$ \\
$\ldots$ \\
S $\rightarrow$ a$_n$ S a$_n$ \\
S $\rightarrow$ $\epsilon$
\end{verse}
Here the final automaton has $3^n$ states. (It records, in effect, one
of three possibilities for each terminal symbol: whether it has not yet
appeared, has appeared and must appear again, or has appeared and need
not appear again.)

There is an important computational improvement that can be made to the
algorithm as described above: instead of removing all the
auxiliary symbols right at the end they can be removed progressively as
soon as they are no longer required; after formulae \ref{e7}--\ref{e8}
have been applied for each non-epsilon rule with dotted rules
$\drt{X,m,*}$, those dotted rules may be removed from the finite-state
language (which typically makes the automaton smaller); and the dotted
rules corresponding to an epsilon production may be removed before
formulae \ref{e7}--\ref{e8} are applied. (To `remove' a symbol means to
substitute it by $\epsilon$: a regular operation.)

With this important improvement the algorithm gives exact
approximations for the left-linear grammars
\begin{verse}
S $\rightarrow$ S a$_1$ \\
$\ldots$ \\
S $\rightarrow$ S a$_n$ \\
S $\rightarrow$ $\epsilon$
\end{verse}
and the right-linear grammars
\begin{verse}
S $\rightarrow$ a$_1$ S \\
$\ldots$ \\
S $\rightarrow$ a$_n$ S \\
S $\rightarrow$ $\epsilon$
\end{verse}
in space bounded by $n$ and time bounded by $n^2$. (It is easiest to
test this empirically with an implementation, though it is also
possible to check the calculations by hand.) Pereira and Wright's
algorithm gives an intermediate unfolded recogniser of size exponential
in $n$ for these right-linear grammars.

There are, however, both left-linear and right-linear grammars for
which the number of states in the final automaton is not bounded by any
polynomial function of the size of the grammar. An examples is:
\begin{verse}
S $\rightarrow$ a$_1$ S\ \ \ S $\rightarrow$ a$_1$ A$_1$ \\
$\ldots$ \\
S $\rightarrow$ a$_n$ S\ \ \ S $\rightarrow$ a$_n$ A$_n$ \\
\underline{A$_1$ $\rightarrow$ a$_1$ X}\ \ \ A$_1$ $\rightarrow$ a$_2$ A$_1$\ $\ldots$\
A$_1$ $\rightarrow$ a$_n$ A$_1$ \\
A$_2$ $\rightarrow$ a$_1$ A$_2$\ \ \ \underline{A$_2$ $\rightarrow$ a$_2$ X}\ $\ldots$\
A$_2$ $\rightarrow$ a$_n$ A$_2$ \\
$\ldots$ \\
A$_n$ $\rightarrow$ a$_1$ A$_n$\ \ \ A$_n$ $\rightarrow$ a$_2$ A$_n$\ $\ldots$\
\underline{A$_n$ $\rightarrow$ a$_n$ X} \\
X $\rightarrow$ $\epsilon$
\end{verse}
Here the grammar has size ${\rm O}(n^2)$ and the final approximation
has $2^{n+1}-1$ states.

Pereira and Wright \shortcite{Pereira96} point out in the context of
their algorithm that a grammar may be decomposed into `strongly
connected' subgrammars, each of which may be approximated separately
and the results composed. The same method can be used with the
finite-state calculus approach: Define the relation $\mathcal{R}$ over
nonterminals of the grammar s.t.\ $A{\mathcal R}B$ iff $B$ appears on
the right-hand side of a production for $A$. Then the relation
${\mathcal S}={\mathcal R}^*\cap({\mathcal R}^*)^{-1}$, the reflexive
transitive closure of $\mathcal{R}$ intersected with its inverse, is an
equivalence relation. A subgrammar consists of all the productions for
nonterminals in one of the equivalence classes of $\mathcal{S}$.
Calculate the approximations for each nonterminal by treating the
nonterminals that belong to other equivalence classes as if they were
terminals. Finally, combine the results from each subgrammar by
starting with the approximation for the start symbol $S$ and
substituting the approximations from the other subgrammars in an order
consistent with the partial ordering that is induced by $\mathcal{R}$
on the subgrammars.

\section{Results with a larger grammar}\label{ressect}

When the algorithm was applied to the 18-rule grammar shown in
figure~\ref{gramfig} it was not possible to complete the
calculations for any ordering of the rules, even with the improvement
mentioned in the previous section, as the automata became too large for
the finite-state calculus on the computer that was being used. (Note
that the grammar forms a single strongly connected component.)
\begin{figure}
\begin{tabbing}
\=
NOM $\rightarrow$ NOM MOD bla\=\kill

\> MOD $\rightarrow$ \>         VP $\rightarrow$ v NP \\
\> MOD $\rightarrow$ p NP \>    VP $\rightarrow$ v S \\
\>  \>                          VP $\rightarrow$ v VP \\
\> NOM $\rightarrow$ a NOM \>   VP $\rightarrow$ v \\
\> NOM $\rightarrow$ n \>       VP $\rightarrow$ VP c VP \\
\> NOM $\rightarrow$ NOM MOD \> VP $\rightarrow$ VP MOD \\
\> NOM $\rightarrow$ NOM S \>   S $\rightarrow$ MOD S \\
\>  \>                          S $\rightarrow$ NP S \\
\> NP $\rightarrow$ \>          S $\rightarrow$ S c S \\
\> NP $\rightarrow$ d NOM \>    S $\rightarrow$ v NP VP \\
\end{tabbing}
\caption{An 18-rule CFG derived from a unification grammar.}
\label{gramfig}
\end{figure}

\begin{figure*}
\begin{verbatim}
% initial approximation:
r( s(s/_/0)^($ *)^s(s/_/z) , a).
% formulae (1)-(2):
r( (#a) - (($ *)-(($ *)^(s(_/_/_)-s(_/_/z))+s))^s(_/_/0)^($ *) , a).
r( (#a) - ($ *)^s(_/_/z)^(($ *)-(s+(s(_/_/_)-s(_/_/0))^($ *))) , a).
% formula (3) for "S -> a S b":
r( (#a) - ($ *)^s(s/1/0)^(($ *)-s(a)^s(s/1/1)^($ *)) , a).
r( (#a) - ($ *)^s(s/1/1)^(($ *)-s(s/_/0)^($ *)) , a).
r( (#a) - ($ *)^s(s/1/2)^(($ *)-s(b)^s(s/1/z)^($ *)) , a).
% formula (4) for "S -> a S b":
r( (#a) - (($ *)-($ *)^s(s/1/0)^s(a))^s(s/1/1)^($ *) , a).
r( (#a) - (($ *)-($ *)^s(s/_/z))^s(vp/2/1)^($ *) , a).
r( (#a) - (($ *)-($ *)^s(s/1/2)^s(b))^s(s/1/z)^($ *) , a).
% formulae (5)-(6) for "S -> ":
r( (#a) - ($ *)^s(s/2/0)^(($ *)-s(s/2/z)^($ *)) , a).
r( (#a) - (($ *)-($ *)^s(s/2/0))^s(s/2/z)^($ *) , a).
% formula (7) for "S -> a S b":
r((#a)-($ *)^s(s/1/0)^(($ *)-(($ -s(s/1/_))*)^(s(s/1/0)+s(s/1/1))^($ *)),a).
r((#a)-($ *)^s(s/1/1)^(($ *)-(($ -s(s/1/_))*)^(s(s/1/0)+s(s/1/2))^($ *)),a).
r((#a)-($ *)^s(s/1/2)^(($ *)-(($ -s(s/1/_))*)^(s(s/1/0)+s(s/1/z))^($ *)),a).
% formula (8) for "S -> a S b":
r((#a)-(($ *)-($ *)^(s(s/1/z)+s(s/1/0))^(($ -s(s/1/_))*))^s(s/1/1)^($ *),a).
r((#a)-(($ *)-($ *)^(s(s/1/z)+s(s/1/1))^(($ -s(s/1/_))*))^s(s/1/2)^($ *),a).
r((#a)-(($ *)-($ *)^(s(s/1/z)+s(s/1/2))^(($ -s(s/1/_))*))^s(s/1/z)^($ *),a).
% define the terminal alphabet:
r( s(s/1/0)+s(s/1/1)+s(s/1/2)+s(s/1/z)+s(s/2/0)+s(s/2/z)+s(a)+s(b), sigma).
% remove the auxiliary symbols to give final result:
r( rem((#a)&((#sigma) *),[_/_/_]) , f).
list_fsa(f).
\end{verbatim}
\caption{The sequence of calculations for approximating S $\rightarrow$
a S b $\mid$ $\epsilon$, coded for the finite-state calculus.}
\label{codefig}
\end{figure*}
\begin{figure*}
\begin{center}\
\psfig{figure=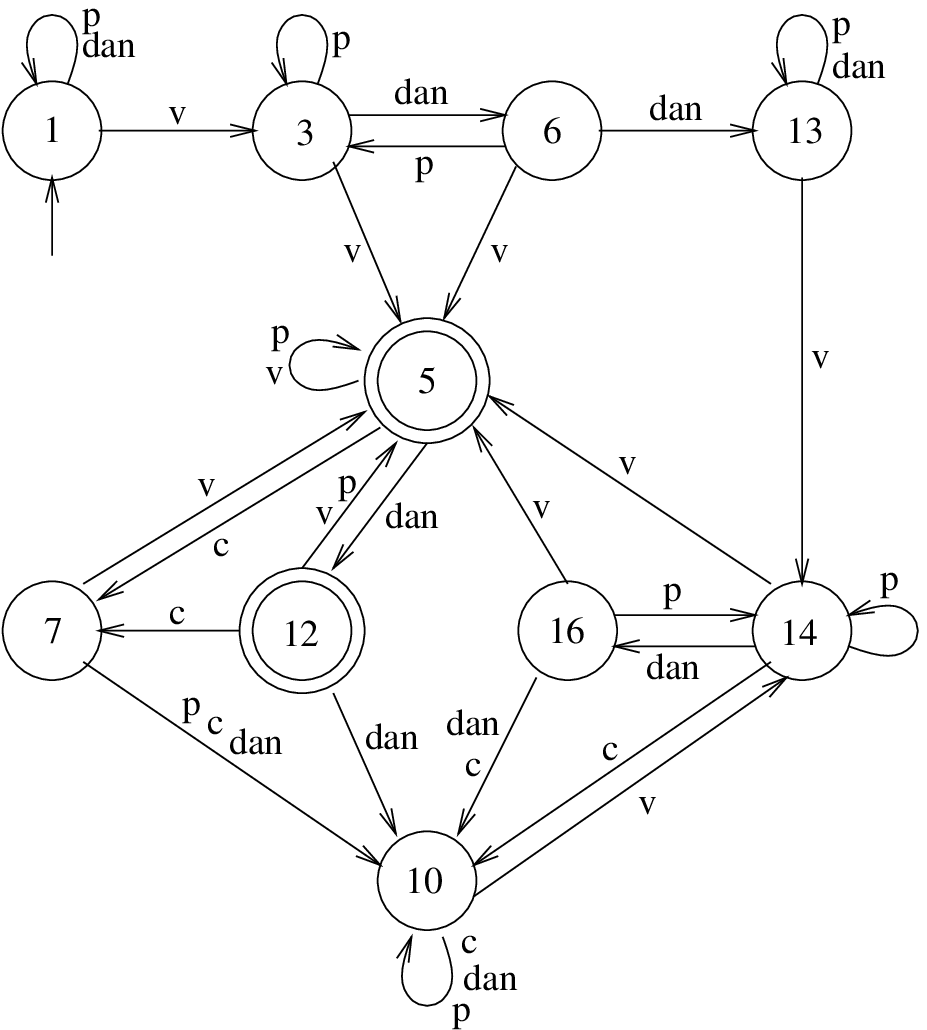,height=7cm}
\hspace{1cm}
\psfig{figure=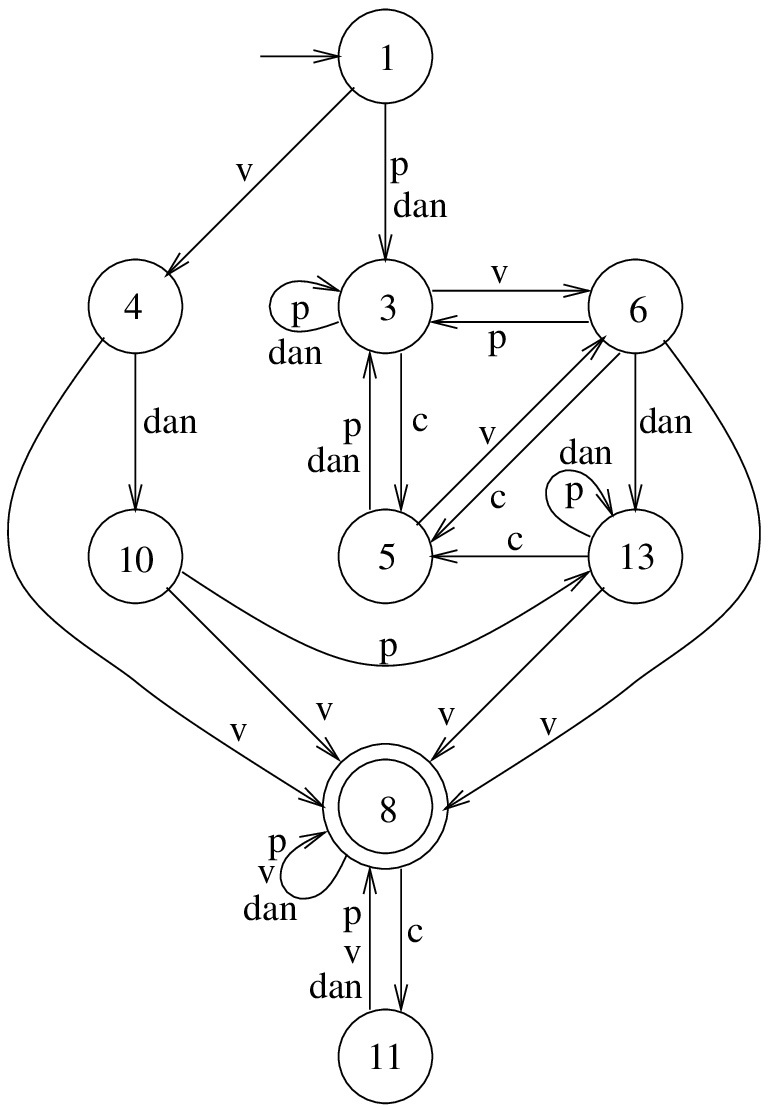,height=7.4725cm}
\ \end{center}
\caption{Finite-state approximations for the grammar in
figure~\ref{gramfig} calculated with the finite-state calculus (left)
and by Pereira and Wright's algorithm (right).}
\label{prettyfig}
\end{figure*}

However, it was found possible to simplify the calculation by omitting
the application of formulae~\ref{e7}--\ref{e8} for some of the
rules. (The auxiliary symbols not involved in those rules could then be
removed before the application of \ref{e7}--\ref{e8}.) In particular,
when restrictions \ref{e7}--\ref{e8} were applied only for the S and VP
rules the calculations could be completed relatively quickly, as the
largest intermediate automaton had only 406 states. Yet the final
result was still a useful approximation with 16 states.

Pereira and Wright's algorithm applied to the same problem gave an
intermediate automaton (the `unfolded recogniser') with 56272 states,
and the final result (after flattening and minimisation) was a
finite-state approximation with 13 states.

The two approximations are shown for
comparison in figure~\ref{prettyfig}. Each has the
property that the symbols \verb|d|, \verb|a| and \verb|n| occur only in
the combination \verb|d| \verb|a|$^*$ \verb|n|. This fact has been used
to simplify the state diagrams by treating this combination as a single
terminal symbol \verb|dan|; hence the approximations are drawn with 10
and 9 states, respectively.

Neither of the approximations is better than the other;
their intersection (with 31 states) is a better approximation than
either. The two approximations have therefore captured different aspects
of the context-free language.

In general it appears that the approximations produced by the present
algorithm tend to respect the necessity for certain constituents to
be present, at whatever point in the string the symbols that `trigger' them
appear, without necessarily insisting on their order, while Pereira and
Wright's approximation tends to take greater account of the
constituents whose appearance is triggered early on in the string: most
of the complexity in Pereira and Wright's approximation of the 18-rule
grammar is concerned with what is possible before the first accepting
state is encountered.

\ignore{ It is also interesting to compare the approximations produced
for the first class of grammars mentioned in section
\ref{sectcompcomp}: whereas the present algorithm gives an
approximation with $3^n$ states in which it is easy to assign a natural
`meaning' to each state Pereira and Wrights's approximation has ?
states and is more difficult to describe in words.}

\section{Comparison with previous work}

Rimon and Herz \shortcite{Herz91,Rimon91} approximate the recognition
capacity of a context-free grammar by extracting `local syntactic
constraints' in the form of the Left or Right Short Context of length
$n$ of a terminal. When $n=1$ this reduces to next(t), the set of
terminals that may follow the terminal t. The effect of filtering with
Rimon and Herz's next(t) is similar to applying conditions
\ref{e1}--\ref{e6} from section \ref{algsect}, but the use of auxiliary
symbols causes two differences which can both be illustrated with the
following grammar:
\begin{verse}
  S $\rightarrow$ a X a $\mid$ b X b\\
  X $\rightarrow$ $\epsilon$
\end{verse}
On the one hand, Rimon and Herz's `next' does not distinguish between
different instances of the same terminal symbol, so any $a$, and not
just the first one, may be followed by another $a$. On the other hand,
Rimon and Herz's `next' looks beyond the empty constituent in a way
that conditions \ref{e1}--\ref{e6} do not, so $ab$ is disallowed. Thus an
approximation based on Rimon and Herz's `next' would be $aa^*+bb^*$,
and an approximation based on conditions \ref{e1}--\ref{e6} would be
$(a+b)(a+b)$. (However, the approximation becomes exact when conditions
\ref{e7}--\ref{e8} are added.)

Both Pereira and Wright \shortcite{Pereira91,Pereira96} and Rood
\shortcite{Rood96} start with the LR(0) characteristic machine, which
they first `unfold' (with respect to `stacks' or `paths', respectively)
and then `flatten'. The characteristic machine is defined in terms of
dotted rules with transitions between them that are analagous to the
conditions implied by formula~\ref{e3} of section \ref{algsect}. When
the machine is flattened, $\epsilon$-transitions are added in a way
that is in effect simulated by conditions \ref{e2} and \ref{e4}.
(Condition \ref{e1} turns out to be implied by conditions
\ref{e2}--\ref{e4}.) It can be shown that the approximation $L_0$
obtained by flattening the characteristic machine (without unfolding
it) is as good as the approximation $L_{\rm 1-6}$ obtained by applying
conditions \ref{e1}--\ref{e6} ($L_0 \subseteq L_{\rm 1-6}$). Moreover, if
no nonterminal for which there is an $\epsilon$-production is used more
than once in the grammar, then $L_0 = L_{\rm 1-6}$. (The grammar in figure
\ref{gramfig} is an example for which $L_0 \not= L_{\rm 1-6}$; the
approximation found in section \ref{ressect} includes strings such as
\verb|vvccvv| which are not accepted by $L_0$ for this grammar.) It can
also be shown that $L_{\rm 1-6}$ is the same as the result of flattening
the characteristic machine for the same grammar modifed so as to fulfil
the afore-mentioned condition by replacing the right-hand side of every
$\epsilon$-production with a new nonterminal for which there is a
single $\epsilon$-production.

However, there does not seem to be a simple correspondence between
conditions \ref{e7}--\ref{e8} and the `unfolding' used by Pereira and
Wright or Rood: even some simple grammars such as `S $\rightarrow$ a S
a $\mid$ b S b $\mid$ $\epsilon$' are approximated differently by
\ref{e1}--\ref{e8} than by Pereira and Wright's and Rood's methods.

\section{Discussion and conclusions}

In the case of some simple examples (such as the grammar `S
$\rightarrow$ a S b $\mid$ $\epsilon$' used earlier) the approximation
algorithm presented in this paper gives the same result as Pereira and
Wright's algorithm. However, in many other cases (such as the grammar `S
$\rightarrow$ a S a $\mid$ b S b $\mid$ $\epsilon$' or the 18-rule
grammar in the previous section) the results are essentially different
and neither of the approximations is better than the other.

The new algorithm does not share the problem of Pereira and Wright's
algorithm that certain right-linear grammars give an intermediate automaton of
exponential size, and it was possible to calculate a useful
approximation fairly rapidly in the case of the 18-rule grammar in the
previous section. However, it is not yet possible to draw general
conclusions about the relative efficiency of the two
procedures. Nevertheless, the new algorithm seems to have the advantage
of being open-ended and adaptable: in the previous section it was
possible to complete a difficult calculation by relaxing the conditions
of formulae~\ref{e7}--\ref{e8}, and it is easy to see how those
conditions might also be strengthened. For example, a more complicated
version of formulae~\ref{e7}--\ref{e8} might check two levels of
recursive application of the same rule rather than just one level and
it might be useful to generalise this to $n$ levels of recursion in a
manner analagous to Rood's \shortcite{Rood96} generalisation of Pereira
and Wright's algorithm.

The algorithm also demonstrates how the general machinery of
a finite-state calculus can be usefully applied as a framework for
expressing and solving problems in natural language processing.

\label{lastpage}

\end{document}